\pdfoutput=1
\documentclass[10pt,twocolumn,letterpaper]{article}

\usepackage{cvis}            

\definecolor{cvisblue}{rgb}{0.21,0.49,0.74}
\usepackage[pagebackref,breaklinks,colorlinks,allcolors=cvisblue]{hyperref}
\hypersetup{pdftitle={Which Way Is Up? Rotated Photos Silently Hurt Multi-View 3D Foundation Models},pdfauthor={Yukun Shi, Minglun Gong}}

\usepackage{enumitem}
\usepackage{capt-of}          
\graphicspath{{figures/}}
\newcommand{\BenchDaDropped}{33}
\newcommand{\BenchMapList}{34}
\newcommand{\ClfCostHigh}{0.06}
\newcommand{\ClfCostLow}{0.03}
\newcommand{\ClfRotHigh}{100\%}
\newcommand{\ClfRotLow}{95\%}
\newcommand{\ClfUpTurnedHigh}{5\%}
\newcommand{\ClfUpTurnedLow}{3\%}
\newcommand{\ConfPhotoMinDa}{70\%}
\newcommand{\ConfPhotoMinMap}{62\%}
\newcommand{\EthRotatedTagUpright}{34}
\newcommand{\FramesUprightN}{407}
\newcommand{\GtFramesHalf}{15\%}
\newcommand{\GtFramesNinety}{97\%}
\newcommand{\GtRecDa}{98\%}
\newcommand{\GtRecVggt}{84\%}
\newcommand{\HolFiles}{812}
\newcommand{\HolPortrait}{155}
\newcommand{\HolPortraitTagged}{40}
\newcommand{\LossDa}{0.36}
\newcommand{\LossMap}{0.18}
\newcommand{\LossOmega}{0.03}
\newcommand{\LossPi}{0.06}
\newcommand{\LossVggt}{0.05}
\newcommand{\OwnLossDa}{0.29}
\newcommand{\OwnLossOmega}{0.04}
\newcommand{\PairMinUpStoredDa}{0.70}
\newcommand{\PairMinUpUpDa}{0.87}
\newcommand{\PairOwnRotStoredDa}{0.02}
\newcommand{\PairOwnRotUpDa}{0.75}
\newcommand{\PairOwnUpStoredDa}{0.73}
\newcommand{\PairOwnUpStoredMap}{0.44}
\newcommand{\PairOwnUpUpDa}{0.71}
\newcommand{\PairOwnUpUpMap}{0.58}
\newcommand{\PortLossHigh}{0.29}
\newcommand{\PortLossLow}{0.04}

\title{Which Way Is Up?\\ Rotated Photos Silently Hurt Multi-View 3D Foundation Models}

\author{Yukun Shi \qquad Minglun Gong\\
University of Guelph\\
{\tt\small \{yshi21, minglun\}@uoguelph.ca}}

\begin{document}
\twocolumn[{%
\renewcommand\twocolumn[1][]{#1}%
\maketitle
\begin{center}
\centering
\includegraphics[width=0.975\linewidth]{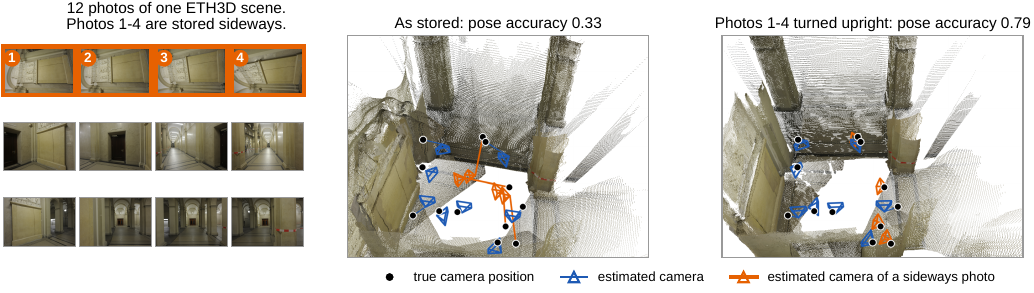}
\captionof{figure}{Four sideways photos are enough to break a reconstruction. Left: twelve photos of one ETH3D scene~\cite{schops2017eth3d}, four of them stored sideways (orange). Middle: with the photos as stored, Depth Anything 3~\cite{lin2025da3} places the cameras of the four sideways photos far from their true positions and splits the walls into misaligned copies. Right: with the same four photos turned upright, the reconstruction is clean. The 3D points and cameras are aligned to the true cameras and seen from above at an angle. Each estimated camera (orange for a sideways photo) is joined by a line to its true position (black dot), so a long line means a large error.}
\label{fig:teaser}
\end{center}

}]
\begin{abstract}
Multi-view 3D foundation models estimate camera poses and 3D points from a set of photos in one forward pass.
Many cameras store a portrait shot sideways, with an orientation tag telling viewers how to turn it, and this tag can be ignored, lost or wrong.
We measure how rotated photos affect five such models on three datasets.
A few rotated photos lower camera pose accuracy for every model, and the sideways photos already in a standard benchmark do the same.
Even the most robust model still loses accuracy.
The failure is also silent: the models' confidence does not reliably flag it, and most official image loaders ignore the tag.
The fixes we test each correct only part of the problem.
These findings call for orientation checks in reconstruction pipelines and benchmarks.
\end{abstract}

\section{Introduction}
\label{sec:intro}

Multi-view 3D foundation models such as VGGT~\cite{wang2025vggt}, $\pi^3$~\cite{wang2025pi3}, Depth Anything 3~\cite{lin2025da3}, MapAnything~\cite{keetha2025mapanything} and VGGT-$\Omega$~\cite{vggtomega2026} estimate camera poses, depth and 3D points from a set of photos in one forward pass.
They need no calibration and no per-scene optimization, which makes them a natural tool for casual capture: a user walks around an object or through a room and takes photos with a phone.

\begin{figure*}[t]
\centering
\includegraphics[width=0.84\linewidth]{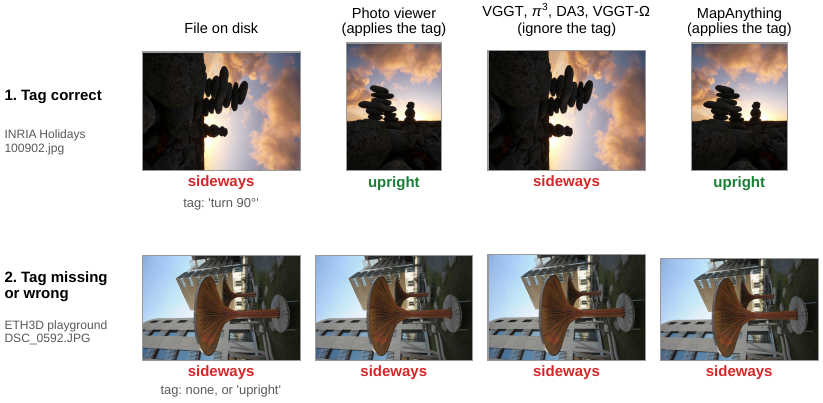}
\caption{What a person sees and what the model gets. Columns: the file on disk, what a photo viewer shows, and what each model's official loader gives the model. Row 1: the tag is correct, so the photo looks upright to the user but reaches four of the five models sideways. Our portrait-first setting simulates this case. Row 2: the tag is missing (ETH3D's benchmark copies) or says ``upright'' (the camera's original file), so every model gets the photo sideways. ETH3D's own sideways photos are this case.}
\label{fig:cases}
\end{figure*}

People also turn the phone for some shots.
We call a photo \emph{rotated} when its stored pixels must be turned to show the scene upright: \emph{sideways} when it is off by 90$^\circ$ and \emph{upside down} when it is off by 180$^\circ$.
We use \emph{portrait} and \emph{landscape} only for the shape of an image.
Many cameras and phones store a portrait shot sideways, as a landscape image, together with an orientation tag in the file's EXIF data~\cite{cipa2026exif} that tells photo viewers how to turn it (\cref{fig:cases}).
A photo therefore reaches a model rotated whenever its tag is ignored, lost or wrong, and all three happen in practice.
Most official image loaders of these models ignore the tag.
ETH3D's~\cite{schops2017eth3d} benchmark images carry no tag, and its original camera files tag every sideways photo as upright.

We measure what rotated photos cost these five models on DTU~\cite{aanaes2016dtu}, ETH3D and Tanks and Temples~\cite{knapitsch2017tanks}, why the problem goes unnoticed, and how well the available fixes work.
We test three kinds of rotated photos.
First, we turn three photos of each set sideways.
Second, we crop the same photos to a portrait shape before turning them, so that each one becomes a wide image with sideways content, which is how a camera stores a portrait shot.
Third, we use the sideways photos that ETH3D already contains.

A few rotated photos lower camera pose accuracy for every model, and the sideways photos already in ETH3D do the same (\cref{fig:teaser}).
The loss is largest for Depth Anything 3 and MapAnything and smallest for VGGT-$\Omega$ and VGGT.
The failure is silent: the models' confidence gives no reliable warning, and a photo stored sideways with a correct tag looks upright in a photo viewer but still reaches four of the five models sideways.
Existing fixes each correct part of the problem.
Reading the tag works only when the tag is present and correct.
Even then, a set with turned portrait shots mixes portrait and landscape photos.
Most official image loaders crop or stretch such a set to one shape, and some models end up no better or even worse.
Single-image orientation classifiers also turn some upright photos, and a gravity estimator~\cite{g3t2026} misses most upside-down photos.

Our contributions are:
\begin{itemize}[leftmargin=*,itemsep=1pt,topsep=2pt]
\item A controlled measurement of what rotated photos cost five multi-view 3D foundation models on three datasets, including the sideways photos already in a standard benchmark.
\item An explanation of why the problem goes unnoticed: what viewers, loaders and benchmark code do with the tag, and what the models' confidence shows.
\item A comparison of the available fixes, with practical advice.
\end{itemize}

\section{Related Work}
\label{sec:related}

\paragraph{Multi-view 3D foundation models.}
DUSt3R~\cite{wang2024dust3r} showed that a network can regress 3D points for a pair of photos without known cameras.
VGGT~\cite{wang2025vggt} extends this to many photos and also predicts cameras and depth in one pass, $\pi^3$~\cite{wang2025pi3} removes the need for a reference photo, Depth Anything 3~\cite{lin2025da3} predicts depth and camera rays, and MapAnything~\cite{keetha2025mapanything} predicts 3D points at metric scale.
VGGT-$\Omega$~\cite{vggtomega2026} simplifies VGGT's architecture so that it can be trained at a much larger scale.

\paragraph{Rotation in other vision tasks.}
Image classifiers lose accuracy when photos are rotated, and data augmentation helps only a little~\cite{engstrom2019spatial}.
Single-image depth networks judge distance mostly from how high an object sits in the image, and they only partly adjust when the camera rolls~\cite{vandijk2019depth}.
Many networks therefore build rotation into their design, so that rotating the input rotates the output~\cite{cohen2016group,weiler2019general}, for example to detect objects in aerial photos at any orientation~\cite{han2021redet}.
Image matching handles rotated photos with rotation-aware keypoint descriptors~\cite{bokman2024steerers}.
We study the multi-view models that replace such matching pipelines, and how they handle rotated photos.

\paragraph{Finding which way is up.}
Orientation classifiers predict the quarter turn that makes a photo upright~\cite{dio2023,iglovikov2020checkorientation}, and gravity estimators predict the camera's roll~\cite{veicht2024geocalib,g3t2026}.
We test both kinds as fixes.

\section{Setup}
\label{sec:setup}

\paragraph{Models.}
We use the released weights of VGGT (1B)~\cite{wang2025vggt}, $\pi^3$~\cite{wang2025pi3}, Depth Anything 3 (DA3-Large-1.1)~\cite{lin2025da3}, MapAnything~\cite{keetha2025mapanything} and VGGT-$\Omega$ (the authors' 416-pixel benchmarking checkpoint)~\cite{vggtomega2026}.

\paragraph{Data.}
We test on the 15 DTU scans~\cite{aanaes2016dtu} and the 6 Tanks and Temples scenes~\cite{knapitsch2017tanks} used by 2DGS~\cite{huang2024twodgs}, and on all 13 high-resolution training scenes of ETH3D~\cite{schops2017eth3d}.
Each set holds 12 photos evenly spaced in capture order.
All photos are stored as landscape images without an orientation tag.
In ETH3D, 34 photos of four scenes (delivery\_area, playground, relief and relief\_2) are stored sideways, turned 90$^\circ$ counter-clockwise, and we draw three sets from each of these scenes, starting at different photos.
All other photos are stored upright.
We find which photos are rotated, and by how much, from the ground-truth cameras alone: the scene's up direction points to the top of an upright photo and to one side of a sideways photo.

\paragraph{Settings.}
We run every set with all photos upright, which is our reference, and with some of its photos rotated.
In the four ETH3D scenes, these are ETH3D's own sideways photos, as stored: one to five per set (about three on average), with three sets per scene.
In the other 30 scenes, we turn three random photos of each set sideways, by 90$^\circ$ clockwise or counter-clockwise.
We make two such random choices per set and report their average.
We also crop the same three photos to a portrait shape before turning them, so that they become wide photos with sideways content, which is how a camera stores a portrait shot (\emph{portrait first}).
Within each setting, the upright and rotated versions use the same photos, including the same crops in portrait first, so the drop in pose accuracy is the cost of the rotation alone.

\paragraph{Canvas.}
Every model sees each photo whole, on a white square of its input size, so a rotated photo gives the model the same image as its upright version, only turned.

\paragraph{Official loaders.}
To test what applying the tag gives in practice (\cref{tab:loaders}), we also run the portrait-first and ETH3D's own sets through each model's official loader, once as stored and once with the sideways photos turned upright, as a correct tag would turn them.
These loaders resize, crop, stretch or pad the photos as they would for any user.

\paragraph{Metric.}
We measure camera pose accuracy with AUC@30, as in VGGT~\cite{wang2025vggt}.
The model estimates the cameras of all twelve photos at once, and the score compares them two at a time, over all 66 pairs.
For each pair, it measures the angular error of the rotation between the two cameras and of the direction from one to the other.
At a given threshold, the pair counts as accurate only if both errors are below it.
AUC@30 is the area under the curve of the fraction of accurate pairs against the threshold, from 0$^\circ$ to 30$^\circ$.
A model that sees a photo turned also sees its camera turned, so before scoring we turn that predicted camera back by the photo's known rotation.

\section{Results}
\label{sec:results}

\subsection{What rotated photos cost}

\begin{table}[t]
\caption{Pose accuracy (AUC@30 $\uparrow$, from 0 to 1, see \cref{sec:setup}). Upright: the 30 scenes whose photos are all stored upright. DTU to All: the accuracy lost when we turn three of their twelve photos sideways. Portrait first: the same three photos, cropped to a portrait shape before turning, against the same crops upright. Last column: the accuracy lost with ETH3D's own sideways photos, in the other 4 scenes, against the same sets turned upright. Every loss is above zero with 95\% confidence (bootstrap over scenes). T\&T: Tanks and Temples.}
\label{tab:cost}
\centering
\small
\setlength{\tabcolsep}{4pt}
\resizebox{\linewidth}{!}{\begin{tabular}{@{}lccccccc@{}}
\toprule
 & & \multicolumn{5}{c}{3 of 12 photos sideways} & \\ \cmidrule(lr){3-7}
Model & Upright & DTU & ETH3D & T\&T & All & Portrait first & ETH3D's own \\
\midrule
VGGT & 0.93 & $-$0.01 & $-$0.10 & $-$0.06 & $-$0.05 & $-$0.04 & $-$0.07 \\
$\pi^3$ & 0.94 & $-$0.01 & $-$0.15 & $-$0.07 & $-$0.06 & $-$0.11 & $-$0.09 \\
Depth Anything 3 & 0.88 & $-$0.29 & $-$0.38 & $-$0.53 & $-$0.36 & $-$0.29 & $-$0.29 \\
MapAnything & 0.77 & $-$0.07 & $-$0.32 & $-$0.24 & $-$0.18 & $-$0.18 & $-$0.29 \\
VGGT-$\Omega$ & 0.96 & $-$0.02 & $-$0.06 & $-$0.02 & $-$0.03 & $-$0.05 & $-$0.04 \\
\bottomrule
\end{tabular}
}
\end{table}

\paragraph{A few rotated photos lower every model's accuracy.}
With three of twelve photos sideways, Depth Anything 3 loses \LossDa{} and MapAnything \LossMap{}.
VGGT and $\pi^3$ lose \LossVggt{} and \LossPi{}, and VGGT-$\Omega$, the most robust model, loses \LossOmega{} (\cref{tab:cost}, column All).
The loss appears on all three datasets.
ETH3D's own sideways photos cost every model too, from \OwnLossOmega{} for VGGT-$\Omega$ to \OwnLossDa{} for Depth Anything 3 and MapAnything.
Every model also loses accuracy in the portrait-first setting (from \PortLossLow{} to \PortLossHigh{}).

\paragraph{The sideways photos' own cameras take most of the damage.}
To see which cameras go wrong, we score two kinds of camera pairs in ETH3D's own sets separately: pairs that include a sideways photo, and pairs of two upright photos (upright pairs).
On ETH3D's own sets, Depth Anything 3's pairs with a sideways photo score \PairOwnRotStoredDa{} while those photos stay sideways and \PairOwnRotUpDa{} once they are turned upright (\cref{fig:teaser} shows one set), but its upright pairs score about the same either way (\PairOwnUpStoredDa{} and \PairOwnUpUpDa{}).
With three sideways photos, its upright pairs also drop, from \PairMinUpUpDa{} to \PairMinUpStoredDa{}, so part of the damage spreads.
For MapAnything, the sideways photos also hurt the cameras of the upright photos in the same set: the upright pairs score \PairOwnUpStoredMap{} while the set's sideways photos stay sideways and \PairOwnUpUpMap{} once they are turned upright.

\subsection{Why the problem goes unnoticed}
\label{sec:unnoticed}

\paragraph{The orientation tag does not reach most models.}
VGGT, $\pi^3$, Depth Anything 3 and VGGT-$\Omega$ read the stored pixels and ignore the tag.
Only MapAnything applies it (checked in each repository on October 14, 2026).
A correctly tagged portrait shot therefore looks right in a photo viewer and reaches four of the five models sideways (\cref{fig:cases}, row 1).
In the first of INRIA Holidays' two archives~\cite{jegou2008holidays}, \HolPortraitTagged{} of the \HolPortrait{} photos that display as portrait are stored as landscape pixels with a tag that turns them.

\paragraph{The tag can be missing or wrong.}
ETH3D's undistorted images, the files that benchmarks use, carry no EXIF data, and in the camera's original files the tag of all \EthRotatedTagUpright{} sideways photos says that they are upright (\cref{fig:cases}, row 2).
A tag can also be wrong the other way: one of the \HolFiles{} Holidays photos is stored upright but carries a tag that turns it, so applying the tag turns a correct photo sideways.

\paragraph{Benchmark code treats these photos as a data problem.}
The code that prepares ETH3D for MapAnything's benchmark turns these \BenchMapList{} photos upright from a list of file names, separately from its loader, and Depth Anything 3's benchmark skips \BenchDaDropped{} of them, so benchmark results do not show what such photos do to a model.

\paragraph{The models' confidence is no reliable warning.}
We average each model's per-pixel confidence over every photo.
For the two models that lose the most, this score separates rotated from upright photos poorly: a rotated photo scores lower than an upright one of the same set in only \ConfPhotoMinDa{} (Depth Anything 3) and \ConfPhotoMinMap{} (MapAnything) of comparisons, where 50\% is a coin flip.

\subsection{What fixes it}

\begin{table}[t]
\caption{Pose accuracy (AUC@30) through each model's official loader. Left of the arrow: the sets as stored, with some photos sideways. Right: the same sets with those photos turned upright, as a correct tag would turn them. Bold: the higher of the two.}
\label{tab:loaders}
\centering
\small
\begin{tabular}{@{}lcc@{}}
\toprule
 & Portrait first & ETH3D's own \\
Model & (30 scenes) & (4 scenes) \\
\midrule
VGGT & 0.87 $\to$ \textbf{0.89} & \textbf{0.74} $\to$ 0.69 \\
$\pi^3$ & \textbf{0.80} $\to$ 0.66 & \textbf{0.81} $\to$ 0.56 \\
Depth Anything 3 & 0.48 $\to$ \textbf{0.75} & 0.39 $\to$ \textbf{0.72} \\
MapAnything & 0.51 $\to$ \textbf{0.61} & 0.33 $\to$ \textbf{0.70} \\
VGGT-$\Omega$ & 0.88 $\to$ \textbf{0.89} & 0.90 $\to$ \textbf{0.92} \\
\bottomrule
\end{tabular}

\end{table}

\paragraph{Reading the tag helps only when the tag is right.}
It fixes row 1 of \cref{fig:cases} but not row 2, and it turns a correct photo sideways when the tag is wrong the other way (\cref{sec:unnoticed}).
Even a correct tag is not enough with the official loaders: four of the five crop or stretch the turned photos to one shape (VGGT-$\Omega$'s only pads them).
Through each model's own loader (\cref{tab:loaders}), turning the sideways photos upright clearly helps Depth Anything 3 and MapAnything, changes VGGT-$\Omega$ little, has mixed effects on VGGT, and hurts $\pi^3$, whose loader stretches every photo to the shape of the first one.

\begin{table}[t]
\caption{Percentage of photos that each fix leaves upright or turns upright, before the photos reach the five tested models. 90$^\circ$: the three sideways photos of a set. 180$^\circ$: upside-down photos, from sets with every photo turned the same way or each photo turned at random. ETH3D's own: the sideways photos in its own sets.}
\label{tab:fixes}
\centering
\small
\setlength{\tabcolsep}{3.5pt}
\resizebox{\linewidth}{!}{\begin{tabular}{@{}lcccc@{}}
\toprule
 & Upright photos & \multicolumn{3}{c}{Rotated photos turned upright} \\ \cmidrule(lr){3-5}
Fix & left upright & 90$^\circ$ & 180$^\circ$ & ETH3D's own \\
\midrule
check\_orientation & 97\% & 98\% & 96\% & 100\% \\
deep-image-orientation-detection & 95\% & 96\% & 95\% & 100\% \\
Gravity estimator (G3T) & 100\% & 97\% & 15\% & 100\% \\
\bottomrule
\end{tabular}
}
\end{table}

\paragraph{Orientation classifiers also rotate upright photos.}
Two released orientation classifiers, check\_orientation~\cite{iglovikov2020checkorientation} (ResNeXt-50) and deep-image-orientation-detection~\cite{dio2023} (EfficientNetV2-S), each run on all four quarter turns of a photo, turn \ClfRotLow{} to \ClfRotHigh{} of the rotated photos upright, but they also turn \ClfUpTurnedLow{} to \ClfUpTurnedHigh{} of the upright photos (\cref{tab:fixes}), which lowers every model's pose accuracy by \ClfCostLow{} to \ClfCostHigh{} on sets with no rotated photo, compared with not using a classifier.

\paragraph{The gravity estimator we test is the safest fix, except for upside-down photos.}
The gravity head of G3T~\cite{g3t2026} turned none of the \FramesUprightN{} photos of the upright sets and every sideways photo in ETH3D's own sets upright.
Of the photos we turn sideways, it turns \GtFramesNinety{} upright (\cref{tab:fixes}), so it recovers \GtRecVggt{} to \GtRecDa{} of their loss, but it turns only \GtFramesHalf{} of the upside-down photos upright.

\paragraph{Practical advice.}
\begin{itemize}[leftmargin=*,itemsep=1pt,topsep=2pt]
\item Apply the orientation tag when it is present.
\item Without a reliable tag, use a gravity estimator, and check separately for upside-down photos, which it mostly misses.
\item After turning photos upright, pad mixed portrait and landscape photos instead of cropping or stretching them.
\item Benchmarks: keep rotated photos in test sets and report them separately.
\end{itemize}

\section{Limitations and Conclusion}
\label{sec:conclusion}

We score camera poses only, on sets of 12 photos, most of them with synthetic rotations.

A few rotated photos lower the pose accuracy of every model we test, nothing in the usual pipeline reliably flags them, and each tested fix covers only part of the problem, so pipelines should check orientation explicitly and benchmarks should keep rotated photos.

{
    \small
    \bibliographystyle{ieeenat_fullname}
    \bibliography{main}
}

\end{document}